\documentclass[conference]{IEEEtran}
\IEEEoverridecommandlockouts

\usepackage{amsmath,amsfonts,bm}

\def\eqref#1{equation~\ref{#1}}

\def\1{\bm{1}}

\DeclareMathAlphabet{\mathsfit}{\encodingdefault}{\sfdefault}{m}{sl}
\SetMathAlphabet{\mathsfit}{bold}{\encodingdefault}{\sfdefault}{bx}{n}

\usepackage{xcolor}
\usepackage{hyperref}
\usepackage{url}
\usepackage{graphicx}
\usepackage{subfigure}
\usepackage{wrapfig}
\usepackage{booktabs}
\usepackage{amsmath}
\usepackage{amsfonts}
\usepackage{amssymb}
\usepackage{mathrsfs}
\usepackage{algorithm}
\usepackage{algorithmic}
\usepackage{accents}

\title{Mastering Strategy Card Game (Hearthstone) with Improved Techniques}

\author{
\IEEEauthorblockN{Changnan Xiao$^*$\thanks{$^*$ Equal contributions, $^{\dag}$ Corresponding author. }}
\IEEEauthorblockA{\textit{ByteDance}\\
changnanxiao@gmail.com}
\\
\IEEEauthorblockN{Qinhan Huang}
\IEEEauthorblockA{\textit{ByteDance}\\
huangqinhan@bytedance.com}
\and
\IEEEauthorblockN{Yongxin Zhang$^*$}
\IEEEauthorblockA{\textit{ByteDance}\\
zhangyongxin.yx@bytedance.com}
\\
\IEEEauthorblockN{Jie Chen}
\IEEEauthorblockA{\textit{ByteDance}\\
chenjiexjtu@gmail.com}
\and
\IEEEauthorblockN{Xuefeng Huang}
\IEEEauthorblockA{\textit{ByteDance}\\
wangfuming@bytedance.com} 
\\
\IEEEauthorblockN{Peng Sun$^{\dag}$}
\IEEEauthorblockA{\textit{ByteDance}\\
pengsun000@gamil.com}
}

\newcommand{\ubar}[1]{\underaccent{\bar}{#1}}

\begin{document}

\maketitle

\begin{abstract}
Strategy card game is a well-known genre that is demanding on the intelligent game-play and can be an ideal test-bench for AI. 
Previous work combines an end-to-end policy function and an optimistic smooth fictitious play, 
which shows promising performances on the strategy card game \emph{Legend of Code and Magic}. 
In this work, we apply such algorithms to \emph{Hearthstone}, 
a famous commercial game that is more complicated in game rules and mechanisms. 
We further propose several improved techniques and consequently achieve significant progress. 
For a machine-vs-human test we invite a \emph{Hearthstone} streamer whose best rank was top 10 of the official league in China region that is estimated to be of millions of players. 
Our models defeat the human player in all \emph{Best-of-5} tournaments of full games (including both deck building and battle), 
showing a strong capability of decision making.
\end{abstract}

\section{Introduction}
\label{sec:intro}

There are many popular strategy card games, 
such as \emph{Hearthstone}, \emph{Gwent}, \emph{Magic: The Gathering}, \emph{Yu-Gi-Oh: Master Duel}, \emph{Marvel Snap}, and so on. 
Most strategy card games include at least two stages, 
where the player constructs a card deck at stage one and battles with the cards-in-deck at stage two. 
The multiple stages increase the complexity of the game, 
which is challenging for the player to make decisions and win the match. 
The multiple stages also open the possibility of intransitive policies (i.e., the Rock-Paper-Scissors phenomenon), 
which characterizes a player's game-play and makes the game intriguing. 
Since strategy card game is a finite two-player zero-sum game, 
the \emph{Nash Equilibrium} exists theoretically if the mixed policy is allowed. 
However, due to the fact that the game is partially observable (imperfect information), 
where the opponent's deck and hand are hidden,  
it's difficult to obtain NE by dynamic programming or other planning-based methods as in conventional perfect information game. 
Due to the gap between NE existence and NE finding, 
strategy card game is thus an appealing test-bench for academic study, 
and it's interesting to achieve the master-level intelligence of the strategy card game.

In this work, we focus on a commercial game, \emph{Hearthstone}. 
Our results demonstrate that the previous algorithm framework \cite{xi2023master}, 
injected with several improved techniques proposed in this study, 
is able to produce master-level agents on this complicated strategy card game.
We evaluate our models in two ways. 
One is machine-vs-machine evaluation. 
Our model has $73.6\%$ winrates versus the baseline, 
where the baseline algorithm won the double championship in COG2022 \emph{Legend of Code and Magic} competition and outperformed other submissions by a large margin. 
We further explore the feasibility to improve the performance by cheating on the opponent deck and achieve $80.2\%$ winrates versus the baseline. 
The other is machine-vs-human evaluation. 
We invite a human player whose best rank was top 10 of the official league in China region that is estimated to be of millions of players. 
Our models of both the normal-version and cheat-version win all the \emph{Best-of-5} (Bo5) tournaments against the human player under the \emph{Conquest} (CQ) rule. 

The rest of this study is organized as follows. 
To begin with, we introduce the background of \emph{Hearthstone}, related works and the approach that won the double championship in COG2022 competition. 
Then we describe how the approach is adapted to \emph{Hearthstone}. 
After that, we show the drawbacks of the approach and how we improve on it. 
Finally, we do the machine-vs-human evaluation to verify the capability of our approach.

\section{Background}
\label{sec:background}

\subsection{Hearthstone}
\label{sec:hearthstone}

\emph{Hearthstone} is a competitive two-player turn-based card game. 
The entire gameplay contains three stages, which are \emph{picking hero} (PH) stage, cards deck building (CB) stage and \emph{battle} (BT) stage. 
The standard game mode is to play with a randomly chosen opponent player. 
The goal is to defeat the opponent hero in the BT stage. 
During the PH stage, the player has to pick a hero, where each hero has a unique Hero Power to be played in the BT stage. 
The opponent's picked hero is hidden to the player until the BT stage (see below).
During the CB stage, the player drafts a deck containing 30 cards, which comes from a card pool consisting of both standard cards and hero-specific cards. 
The deck is hidden information to the opponent.
During the BT stage, in each turn, the player draws some hand cards from the deck and plays cards under limited resource constraint, i.e., the mana that gradually increases with the number of turns. 
The player can play multiple cards or execute multiple activities under the resource constraint. 
The player can pre-end the turn or exhaust all resources. 
Once the player's turn ends, it becomes the opponent's turn to play. 
The goal is to defeat the opponent hero by reducing its health point to zero. 

Our game environment is modified from the open-sourced implementation Hearthbreaker\footnote{https://github.com/danielyule/hearthbreaker}.
Our code supports three heroes: Mage, Warrior and Hunter. 
A card pool of 350+ cards in total are developed, 
where on average 270+ cards, 
including 240 common cards and 30+ hero-specific cards, 
are available to each hero during the CB stage. 
The game rules and mechanisms are approximately equivalent to the commercial \emph{Hearthstone} game of April 2015 (the Blackrock Mountain version).  

There exists an official ranking system of Standard Mode to measure the gameplay level of all players in a particular region\footnote{https://hearthstone.fandom.com/wiki/Ranked}.
By the end of year 2022, there are globally four geographically divided regions: 
Americas, Europe, Asia-Pacific and China\footnote{https://hearthstone.fandom.com/wiki/Region}.
Except for New Player Ranks, there are Normal Ranks and Legend Ranks. In Normal Ranks, there are 5 leagues (Bronze, Silver, Gold, Platinum, Diamond) with ranks from 10 to 1. 
Beyond Diamond 1, it's the Legend Ranks. 
The icon of Legend Ranks shows a number, $n$, for each legendary player, 
which represents that the player is the best $n$-th player in the region. 

Besides the ranking system, 
tournament is a more straightforward way to evaluate a player's level, 
which is the main game format in typical eSport activities. 
We adopt the \emph{Best of 5} (Bo5) tournament under \emph{Conquest} (CQ) \footnote{Rules can be found at https://hearthstone.fandom.com/wiki/Tournaments.} rule. 
In a tournament, each player has to prepare multiple decks and each hero can only have at most one deck. 
CQ requires that any hero (and the deck it brings) that wins a match cannot be played in the remaining matches. 

\subsection{Related Work}
\label{sec:related_work}

\emph{Deep Reinforcement Learning} (DRL) has shown many impressive results on different kinds of games, including single-agent video games \cite{badia2020agent57,wayne2018unsupervised}, two-player games \cite{silver2016mastering} and also multi-player games \cite{berner2019dota,vinyals2019grandmaster,li2020suphx}. 
Strategy card games is a genre that has the following features: \emph{Partially-Observable Markov Decision Process} (POMDP) and \emph{two-player zero-sum} (2p0s). 

To solve a MDP, the objective of DRL is to 
\begin{equation}
    \text{maximize}\ \mathbb{E}_{\pi} [\sum_t \gamma^t r_t],
\label{eq:rl_obj}
\end{equation}
where the expectation is over all trajectories sampled by interactions between the environment and the target policy $\pi$, $t$ is the timestamp, $\gamma$ is the discount rate and $r_t$ is the reward function. 
To deal with POMDP, various methods are proposed in the literature: 
to enlarge the observation space \cite{kapturowski2019recurrent,wayne2018unsupervised,parisotto2020stabilizing}, 
to learn a latent dynamic of the state \cite{hafner2019learning}, 
or to adopt a centralized critic \cite{foerster2018counterfactual}. 

A 2p0s game is fundamentally different from a single-player MDP. 
To solve 2p0s game with RL, a meta strategy is required as wrapper for RL in order to efficiently find \emph{Nash Equilibrium} (NE). 
Regret minimization are widely studied to solve the imperfect information games in a value-centric manner \cite{zinkevich2007regret,brown2019deep}. 
The Smooth Fictitious Play (SFP) \cite{mertikopoulos2019learning}, which is essentially Dual Averaging (DA) method \cite{nesterov2009primal}), is yet another promising way to find NE in a policy-centric manner. 
However, it's somewhat inconvenient in practice, as SFP/DA is only guaranteed to be \emph{average-iterate} convergence (which means the mixture of the policy checkpoints during training converges), 
and it is notoriously difficult to learn and maintain the average policy when using neural network as policy function approximator in a multi-step game \cite{heinrich2015fictitious}. 
Some recent works propose a NE finding algorithm with \emph{last-iterate} convergence, 
where it is unnecessary to separately learn an average policy 
and the policy of the last iteration suffices to be optimal with some constraints \cite{lee2021last,abe2022last,sokota2022unified}. 
In particular, \cite{xi2023master} modifies the standard SFP and propose the Optimistic SFP that also shows last-iterate convergence. 

Hearthstone AI was studied in previous literature, 
where tree search based methods \cite{santos2017monte,zhang2017improving,zolboot2022hearthstone} and evolutionary methods \cite{garcia2020optimizing} are proposed, 
which are further combined with Supervised Learning and Reinforcement Learning \cite{swiechowski2018improving}. 
In this work, however, we solely rely on deep reinforcement learning and fictitious play which are trained from scratch and do not require look-ahead search in both training and inference-time. 

Meanwhile, to our best knowledge, there were no previous Hearthstone AIs showing the ability to defeat top human players in full-game. 
In \cite{zolboot2022hearthstone} and the \emph{hearthstone-ai project}\footnote{https://github.com/peter1591/hearthstone-ai}, 
there were qualitative descriptions of the AIs being competitive with human players, 
but the level of the human players and the quantitative evaluation are absent.
In \cite{swiechowski2018improving} human-vs-machine results were reported, 
but the evaluation was fixed on two pre-defined decks and the AI seemed to achieve a $< 50\%$ winrate against the Legendary level human players (Table III of \cite{swiechowski2018improving}). 
While the general Legendary level players are top $0.5\%$, 
the exact rank of the human players were not reported in \cite{swiechowski2018improving}. 
In this work, 
the evaluation is in full game where the AI and the human player have the complete freedom in both deck building and battle. 
Our AI can defeat a top 10 human player in the Legendary league, 
which is estimated to be at least top $0.0005\%$ human players.

There are also Hearthstone AI competitions \cite{Dockhorn2019}, 
but no human players are involved as the benchmark.

\subsection{End-to-end policy and Optimistic Smooth Fictitious play}
\label{sec:e2e_and_osfp}

\emph{Legends of Code and Magic} (LoCM) is a similar strategy card game that is used for research purpose. 
Previous researches on LoCM mostly use the alternating training: When training the CB policy, BT policy is fixed. 
When training the BT policy, CB policy is fixed. 
The CB policy is optimized by algorithms such as evolutionary method\cite{yang2021deck}, where the score function of cards is defined by expert rules and data mining. 
The BT policy is optimized by search based methods such as \emph{Monte-Carlo Tree Search} (MCTS) \cite{browne2012survey} or reinforcement learning \cite{vieira2019reinforcement}. 

\cite{xi2023master} overcomes the weaknesses of alternating training and propose an \emph{end-to-end} (E2E) policy function. 
It combines the E2E policy function and an \emph{Optimistic Smooth Fictitious Play} (OSFP). 
The resultant algorithm won the double championship of LoCM competitions of COG2022. 

The E2E policy is a single policy function to estimate CB policy and BT policy. 
Denote the observation as $o$ and the indicator of stage as 
$$
\delta = \left\{\begin{aligned}1,\,\text{at CB stage},\\ 0,\,\text{at BT stage}. \end{aligned} \right.
$$
The E2E policy is calculated by 
\begin{equation}
    \pi_\theta(\cdot|o) = \delta  \pi_{\theta_{CB}}(\cdot|o) + (1 - \delta) \pi_{\theta_{BT}}(\cdot|o),
\label{eq:e2e}
\end{equation}
where $\theta_{CB}$ and $\theta_{BT}$ are parameters of CB policy and BT policy respectively. 
There may exist overlapping parameters between $\theta_{CB}$ and $\theta_{BT}$. 
The value function is estimated by extracted features from both CB policy and BT policy. 

Like the traditional Smooth Fictitious Play \cite{fudenberg1998learning,leslie2006generalised}, 
OSFP is an iterative algorithm that the current policy is updated by playing against the mixture of historical models as the opponent, a procedure called Smooth Best Response (SBR). 
However, 
OSFP puts higher mixing weights for the latest model, 
which leads to the property of last-iterate convergence \cite{xi2023master}. 
It means that the training can be stopped anytime, 
and it suffices to simply deploy the last checkpoint model which is guaranteed to be the strongest one so far. 
Finally, like the Deep Fictitious Play in recent literature\cite{heinrich2016deep}, 
the SBR is solved by RL method where the parameters of the policy net and value net are updated.

\section{Application on Hearthstone}
\label{sec:application}

We arrange the remaining contents in the following order. 
In this section, we introduce the beginning of our work, which is our initial baseline. 
In the next section, we introduce the problems we met and the techniques to overcome the difficulties. 

\emph{Hearthstone} contains three stages, which are \emph{picking hero} (PH) stage, \emph{cards deck building} (CB) stage and \emph{battle} (BT) stage. 
The PH stage is simply choosing a hero. Since PH is a new stage compared to LoCM and previous work \cite{xi2023master}, it's a natural question whether we should include PH into our E2E policy function. 
The conclusion is \textbf{NO} and further discussions are shown later. 
The observation space of CB stage contains information of all cards, the selected cards and the cards that can be selected now. 
The action space of CB stage is a categorical distribution to select a card. 
The observation space and the action space of BT stage is more complicated than that of previous work \cite{xi2023master}. 
All observable information of a human player is included in the observation space, which is listed in Tab. \ref{tab:obs} and Fig. \ref{fig:nn_structure}. 
The action space of each step is a single categorical distribution. 
Playing one card in the game needs two operations, 
which is represented by a binary tuple in the form of \emph{(type, target)}. 
\emph{Type} is one of \emph{\{my hand card, my board card, opponent's board card, my hero power card}\footnote{We regard \emph{hero power} as a card that can be selected and applied.} \emph{and end turn card}\footnote{We regard \emph{end turn} as a card that works for ending the turn.}\emph{\}}. 
\emph{Target} is one of \emph{\{my hero, opponent's hero, my board card and opponent's board card\}}. 
For instance, the first operation is to select a card as \emph{type} and the decision type is \emph{selecting}. 
The second operation is selecting a card as \emph{target}. 
Depending on the selected card of the first operation, the decision type of the second operation may differ. 
We process the logic and apply the \emph{type} card onto the \emph{target} card, which completes a play of a card. 

By such an auto-regressive action space decomposition \cite{vinyals2019grandmaster},
the action size is at most several tens at each time step. 
Also, we employ a 0/1 \emph{action mask} to indicate the available actions that varies at each time step. 
When computing the softmax policy inside the neural net,
the probabilities of those unavailable actions are zeroed using this action mask.
A complete list of observation space and action space of CB stage and BT stage are shown in Appendix \ref{app:obs_and_act}. 

The function approximation follows the E2E policy function \cite{xi2023master}. 
The policy function is constructed by \eqref{eq:e2e}, where $\pi_{\theta_{CB}}$ and $\pi_{\theta_{BT}}$ are estimated respectively, where ${\theta_{CB}}$ and ${\theta_{BT}}$ share card embedding and hero embedding. 
The value function is estimated by intermediate features of $\pi_{\theta_{CB}}$ and $\pi_{\theta_{BT}}$. 
The entire network is shown in Appendix \ref{app:network}. 

The policy gradient that we have applied is a combination of V-Trace \cite{espeholt2018impala} and UPGO \cite{vinyals2019grandmaster}. 
For notational convenience, 
at step $t$ we denote by $V_t$ the value function predicted by the neural net, 
by 
\begin{equation*}
\begin{aligned}
    \delta_t &= r_t + \gamma V_{t+1} - V_{t}, \\
    \rho_t   &= \pi(a_t|o_t)/\mu(a_t|o_t)
\end{aligned}
\end{equation*}
the temporal difference and the importance sampling weight, respectively. 
$\pi$ is learning policy; $\mu$ is behaviour policy that generates the data, which is usually $\pi$ in previous iteration.
V-Trace estimates the value function by off-policy correction:
\begin{equation}
    v_t = V_t + \sum_{i=0}^k \gamma^i [\prod_{j=0}^{i-1}\min(\rho_{t+j}, \Bar{c})] \min(\rho_{t+i}, \Bar{\rho}) \delta_{t+i}, 
\label{eq:vtrace}
\end{equation}
where $\Bar{c}$ and $\Bar{\rho}$ are hyperparameters that reduce the variance but introduce a bias. 
When $\Bar{c} \leq \Bar{\rho}$, convergence of \eqref{eq:vtrace} is guaranteed. 
The policy gradient of V-Trace is given by 
\begin{equation}
    \min(\rho_{t}, \Bar{\rho}) (r_t + \gamma v_{t+1} - V_t) \nabla_\theta \log \pi_\theta(a_t|o_t). 
\label{eq:pg_vtrace}
\end{equation}
The value function is trained by minimizing $l2$ distance between $V_t$ and $v_t$. 

Since \emph{Hearthstone} is a 2p0s game, 
a meta algorithm is required as the wrapper of DRL in order to find NE solution. 
We follow previous work on LoCM and apply OSFP \cite{xi2023master}. 
We provide the pseudo code of OSFP in Appendix \ref{app:osfp}.

\section{Improvements and Experiments}
\label{sec:improvements_and_experiments}

\subsection{Machine-vs-Machine Evaluation Criterion}
\label{sec:eval_crit}

To evaluate two models $A$-vs-$B$, 
we run $2 \times 3 \times 3 \times 2500$ matches and calculate the average winrate. 
Both $A$ and $B$ play three heroes (henceforth $3 \times 3 \times$). 
Each model can play on first turn at BT stage or not (henceforth $2 \times$). 
For each condition, we run $2500$ matches. 
We say $A$ increases $a\%$ w.r.t. $B$ when the winrate of $A$-vs-$B$ is $(50+a)\%$. 
All models are compared after training for 2 days, unless otherwise specified. 
Tab. \ref{tab:baselines} gives the results, 
where each model is abbreviated as ``letter digit'' that stands for the type and the version number, 
e.g., ``b4'', ``c5''.
We also use ``letter digit how-many-days'' to indicate how long the corresponding model trains, 
e.g., ``b4-23day'' means the b4 model checkpoint trained for 23 days.
See the running texts in the following contents.

\begin{table}[h]
\begin{center}
\scalebox{1.0}{
\begin{tabular}{c|cccccc}
\toprule
(\%) & b0 & b1.5 & b2 & b3 & b4 & c5 \\
\midrule
b0 & 50.0 & 43.0 & 37.2 & 24.3 & 26.4 & 19.8 \\
b1.5 & 57.0 & 50.0 & 40.0 & 30.4 & 29.7 & 21.3 \\
b2 & 62.8 & 60.0 & 50.0 & 34.9 & 38.3 & 29.9 \\
b3 & 75.7 & 69.6 & 65.1 & 50.0 & 43.5 & 39.0 \\
b4 & 73.6 & 70.3 & 61.7 & 56.5 & 50.0 & 44.5 \\
c5 & 80.2 & 78.7 & 70.1 & 61.0 & 55.5 & 50.0 \\
\bottomrule
\end{tabular}
}
\caption{Winrate table of all models. The entity at $(i, j)$ represents the winrate of ($i$-th row model)-vs-($j$-th column model). All models are described from Sec.\ref{sec:exclude_ph} to Sec.\ref{sec:cheat}. }
\label{tab:baselines}
\end{center}
\end{table}

\subsection{Exclusion of PH from E2E}
\label{sec:exclude_ph}

Previous work on LoCM only considers CB stage and BT stage. 
The E2E policy for the two stages is define as in \eqref{eq:e2e}. 
However, \emph{Hearthstone} has three stages. 
To deal with the additional PH stage, there exist two options. 
One option is to simply sample a hero for each side of a match. The distribution is the prior distribution of heroes, and we simply let it to be a uniform distribution. 
The other option is including PH into the E2E policy function. 
To extend E2E into three stages, we define 
$$
\delta_{X} = \left\{
\begin{aligned}
&1,\ \text{if at $X$ stage}, \\
&0,\ \text{else.} \\
\end{aligned}
\right.
$$
Then the E2E policy can be defined by 
\begin{equation}
\pi_\theta(\cdot|o) = \delta_{PH} \pi_{\theta_{PH}}(\cdot|o) + \delta_{CB} \pi_{\theta_{CB}}(\cdot|o) + \delta_{BT} \pi_{\theta_{BT}}(\cdot|o). 
\end{equation}
$\pi_{\theta_{PH}}$ is a single-step policy with ``null'' observation (just like a Bandit), 
where each action represents the probability of picking the hero. 

\begin{figure}[ht]
\centering
\includegraphics[width=0.96\linewidth]{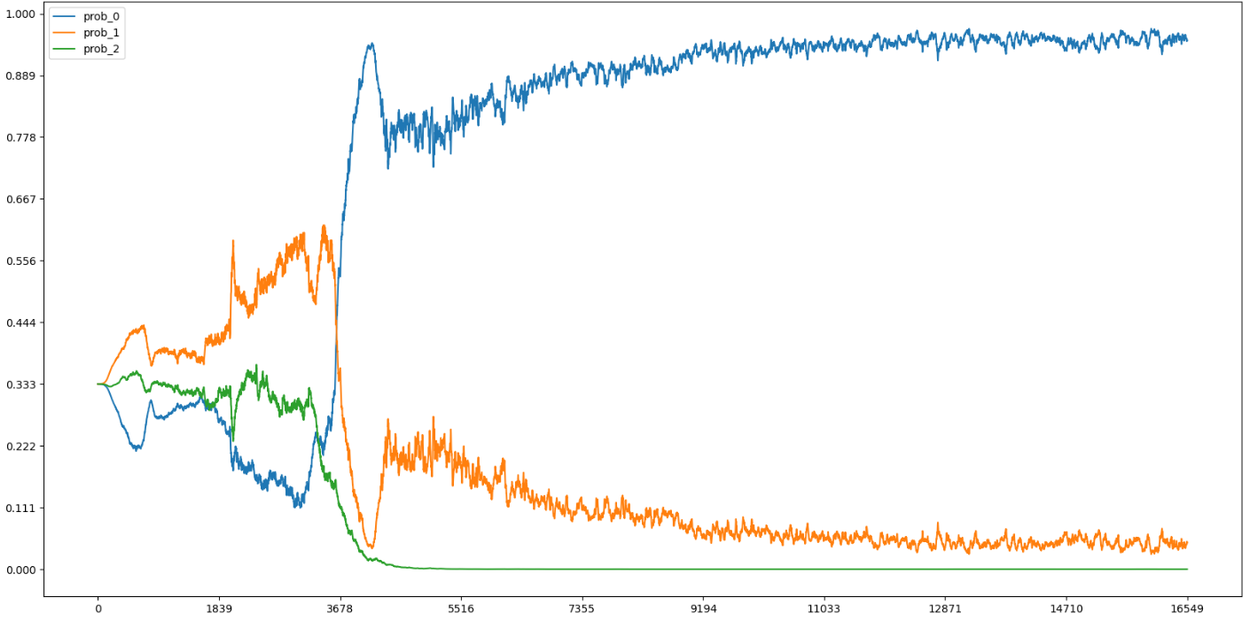}
\caption{Distribution of picked hero during training process when including PH into E2E. \emph{Blue} for hunter, \emph{orange} for mage and \emph{green} for warrior. }
\label{fig:ph}
\end{figure}

We expect that the distribution of picked heroes is similar to the players' on the Blackrock Mountain version of the game when we include PH into E2E policy. 
However, our experiment fails to do so. 
Fig. \ref{fig:ph} shows the distribution of picked hero and it obviously converges to a single hero. 
We have tuned the weight of entropy penalty of $\pi_{\theta_{PH}}$, but it converges to either a single hero or the uniform distribution. 
The policy function only performs well on hunter vs hunter, but performs poor on all the other matches. 
Hence, we exclude PH from E2E and simply sample a hero from the uniform distribution. 
We denote the baseline that uniformly samples a hero and is trained with E2E policy as \emph{baseline-0} (b0). 
Notice that this beginning baseline is actually strong, because the same routine has already won the double championship on LoCM \cite{xi2023master}. 

\subsection{Improvement on $\gamma$}
\label{sec:improve_gamma}

Previous work on LoCM \cite{xi2023master} trains E2E policy with discount rate $\gamma = 0.99$.
In strategy card games, however, each episode has limited steps $(\sim 100)$, 
and a non-zero reward $(+1/-1)$ indicating game win/loss is only given at the end of the episode. 
Hence, the episodic length is limited and the variance of return is bounded.
In this setting,
$\gamma = 1.0$ should feasibly and faithfully recover the return we are interested in (i.e., the game win/loss), 
while $\gamma = 0.99$ fails to do so, despite $\gamma = 0.99$ typically leads to faster training convergence.
Considering this, we set $\gamma = 1.0$ instead of $\gamma = 0.99$,
and find out that the winrate increases around $7\%$ w.r.t. b0. 
We denote b0 with $\gamma = 1.0$ as \emph{baseline-1} (b1). 
Note in Tab. \ref{tab:baselines} we do not include b1, as it is shortly replaced by b1.5, see below section.

\subsection{Random Initialization via Random-CB}
\label{sec:random_cb}

Intuitively, a diversified deck helps the learning of BT policy. 
To encourage such explorations, 
we implement random initialization during the CB stage, which is called random-CB. 
Before each match, we sample a random-CB number $n \in \{0, 1, 2, 4\}$ for each player with probability $(0.5, 0.25, 0.125, 0.125)$. 
Once $n$ is chosen, the first $n$ steps of CB stage choose cards by uniformly random sampling. 
This is achieved by multiplying $0$ on the logits of $\pi_{\theta_{CB}}$ for the first $n$ steps, shown in Fig. \ref{fig:nn_structure}. 
When evaluating the model, we disable random-CB. 
We find the model is more prone to build a ``human-like'' deck with random-CB.
For example, we observed one of the decks built by the model is much like the so called ``Flamewaker Mage'' as was invented by and popular in the community back in 2015\footnote{http://games.sina.com.cn/o/z/hs/2015-04-28/1127616179.shtml}. 
Even though the winrates are on-par with b1, we regard it as progress. 
We denote b1 with random-CB as \emph{baseline-1.5} (b1.5). 

\subsection{Latent Off-Policy Reduction via Balancing}
\label{sec:balance}

Since a large scale training system is asynchronous, off-policy problem is inevitable. 
However, we can balance the data production (i.e., rollout during the environment-agents interaction) and consumption (i.e., learning data batch to update Neural Net) to improve the performance. 
We denote the producing speed as $s_p$ and the consuming speed as $s_c$. 
We usually require $s_p \geq s_c$ to fully exploit the utility of GPU resource. 
We denote $s_p = c \cdot s_c$ with $c \geq 1$. 
Since it's almost impossible to let $c = 1$ in exact, we only consider $c > 1$. 

In a distributed training system that adopts the Actor-Learner architecture \cite{espeholt2018impala,sun2020tleague}, 
there are two options to deal with data overproduction.
The first option is a \emph{ring} buffer. 
The Learner samples uniformly from the buffer. 
When overproduction happens, 
the Actor directly overwrites the buffer in a ring order, 
disregarding whether the existed data have been consumed or not. 
The second option is a \emph{queue} buffer. 
The Learner takes the first-in-first-out samples from the buffer. 
When overproduction happens, the Actor blocks and puts data into the buffer until the buffer is not full. 
Because the \emph{ring} buffer cannot control how many times each data is used exactly, 
it involves additional sampling error compared with the desired stationary distribution. 
Hence, we adopt the \emph{queue} buffer. 
Since the \emph{queue} buffer requires the Actor to wait, we expect $c$ to be close to $1$, where the off-policyness is alleviated. 
We find that b1.5 has $c > 2$, therefore we reduce the Actors by half. 
This surprisingly increases $10.0\%$ winrates w.r.t. b1.5. 
We denote b1.5 with only half Actors $(c \sim 1)$ as \emph{baseline-2} (b2). 

\subsection{Improvement on V-Trace}
\label{sec:improve_vtrace}

During the training process, we find the convergence speed is not satisfactory. 
If we look at V-Trace in \eqref{eq:vtrace}, it only requires $\Bar{c} \leq \Bar{\rho}$ for convergence rather than $\Bar{c} \leq \Bar{\rho} \leq 1$. 
Many previous works take $\Bar{\rho} = 1$ \cite{espeholt2018impala,xi2023master}, but it's not necessary. 
Hence, we increase the hyperparameter $\Bar{\rho}$ to increase the variance and reduce the bias. 
We also find many $\rho$s from b2 to be extremely small $(\sim 10^{-4} \text{ or lower})$, which eliminates the trace. 
Hence, we also clip $\rho$ to be larger than a threshold to prevent diminish of the trace. 
Define $$\text{clip}(x, \alpha, \beta) = \min(\max(x, \alpha), \beta).$$ 
We revise the V-Trace and estimate the value function by 
\begin{equation}
    v_t = V_t + \sum_{i=0}^k \gamma^i [\prod_{j=0}^{i-1}\text{clip}(\rho_{t+j}, \ubar{c}, \Bar{c})] \text{clip}(\rho_{t+i}, \ubar{\rho}, \Bar{\rho}) \delta_{t+i}, 
\label{eq:vtrace2}
\end{equation}
where we use $\ubar{c} = \ubar{\rho} = 0.001$ and $\Bar{c} = \Bar{\rho} = 1.007$ in practice. 

We also find the policy gradient given by clipped importance sampling in \eqref{eq:pg_vtrace} is not stable. 
Hence, we replace it by that of PPO \cite{schulman2017proximal}, which is 
\begin{equation}
\begin{aligned}
    \nabla_\theta \min (& (r_t + \gamma v_{t+1} - V_t)\frac{ \pi_\theta(a_t|o_t)}{\mu(a_t|o_t)}, \\
                        & (r_t + \gamma v_{t+1} - V_t)\text{clip}(\frac{ \pi_\theta(a_t|o_t)}{\mu(a_t|o_t)}, 1-\epsilon, 1+\epsilon) ),
\label{eq:ppo}
\end{aligned}
\end{equation}
where we use $\epsilon = 0.2$ in practice. 
Note in \eqref{eq:ppo} the value estimate $v_{t+1}$ is still V-Trace as in \eqref{eq:vtrace}, 
which differs from conventional PPO implementation.

With these improvements on V-Trace, the winrate increases $15.1\%$ w.r.t. b2. 
We denote b2 with improvements on V-Trace as \emph{baseline-3} (b3). 

\subsection{Model Isolation by Hero}
\label{sec:iso_hero}

All baselines so far are trained with 8 GPUs accompanied with CPUs that ensures data are properly produced and consumed. 
It's a natural question that whether we can achieve better performance with more computational resources. 
Unfortunately, when we limit the capacity of neural network, we find that the performance wouldn't be better when we double/triple the batch size/GPUs/CPUs. 
Considering the fact that each hero has a unique play style, we isolate the model for each hero. 
This triples the computational resources, but it also promotes the performance. 
The hero isolation increases $6.5\%$ winrates w.r.t. b3. 
We denote b3 with hero isolation as \emph{baseline-4} (b4). 

\subsection{Cheat}
\label{sec:cheat}

Up to now, b4 is strong enough to play with the top players, shown later. 
Noticing the fact that \emph{Hearthstone} is a POMDP, we wonder whether cheating by peeking the hidden information leads to a stronger policy. 
We include the opponent's first $n$ chosen cards in deck into the observation spaces of CB stage and BT stage (Note the opponent hand cards are still unobservable during BT stage.). 
Hence, the cheating policy is expected to build adversarial deck and to play adversarial strategy against the opponent. 
However, if both the target policy and the opponent's policy in a game cheat during the training process, it's inconsistent when the target policy plays with the no cheating policies. 

To mitigate the gap, we use an asymmetric training strategy. 
At the beginning of each game, we sample $n_1, n_2 \sim \text{Uniform}(\{0, 1, \dots, 30\})$. 
Then we let $n_2 = \min(n_1, n_2)$. 
Let the target policy, which sends training data to Learner, to observe the first $n_1$ chosen cards of the opponent's deck. 
Let the opponent's policy, which doesn't send training data to Learner, to observe the first $n_2$ chosen cards of the opponent's deck. 
In this way, we guarantee the target policy can always observe more information than the opponent's. 
When evaluating the cheating policy, we sample $n \sim \text{Uniform}(\{0, 1, \dots, 30\})$ and let the cheating policy to observe the first $n$ chosen cards of the opponent's deck. 
It increases $5.5\%$ winrates w.r.t. b4. 
We denote it as \emph{cheat-5} (c5). 

\section{Machine-vs-Human Evaluation}
\label{sec:machine_vs_human}

We develop a dedicated web GUI for machine-vs-human test, 
see Appendix \ref{app:gui}.

By the end of November 2022, 
our models (some intermediate versions between b3 and b4, not reported in Tab. \ref{tab:baselines}) were able to stably defeat two legendary-level human players who best ranked around 20000 and 1000, respectively, in the official league of China region.
In December 2022, we invited another much stronger human player for the machine-vs-human test. 
The human player is a \emph{Hearthstone} streamer, 
whose best rank was top 10 of the official League in China region which is estimated to be of millions of players. 
The human player ranked 60 to 70 when participating in the test.
Before the serious tournament, we deployed a b4-2day model for the human player to play around and warm up. 
Such a casual playing is designed to help the human player getting familiar with the Blackrock Mountain version of the game, 
which is different from the latest commercial \emph{Hearthstone} version. 
The human player can also build and save decks that are used in the later serious tournaments.
The human player can play warm-up matches as many times as wanted. 
We have told the human player that the official agent is much stronger than the warm-up agent. 
Once the human player notified us that he was ready, we started four official CQ tournaments. 
We deployed two different models, b4-23day (the best normal-version) and c5-16day (the best cheat-version that peeks the deck). 
Each model played two tournaments. 
The model picks hero by uniformly sampling, when it is required to change hero by the rule. 
All tournaments are Bo5. 
The human player doesn't know that we deploy different models. 
We also set monetary award for the evaluation: 
For each formal match played, the human player earns xx bucks; 
But if she/he wins that match, 5 times bucks are earned.

\begin{table}[h]
\begin{center}
\scalebox{1.0}{
\begin{tabular}{c|cccccc}
\toprule
 & tournament 1 & tournament 2 \\
\midrule
b4-23day-vs-human & 3:0 & 3:0 \\
c5-16day-vs-human & 3:1 & 3:2 \\
\bottomrule
\end{tabular}
}
\caption{Machine-vs-human evaluation result. }
\label{tab:machine-vs-human}
\end{center}
\end{table}

The result is shown in Tab. \ref{tab:machine-vs-human}. 
It's surprising that the models win all the four tournaments. 
According to the online interview of the human player after the test and by reviewing the game replay, we have the following conclusions. 
\emph{Firstly}, the E2E policy and the OSFP training can yield very strong card game agent. 
The human player says that (English translation from Chinese, the same below): ``Generally speaking, the AI's performance is not bad. It knows when to clear the minions. The one turn kill is also accurate.'' 
\emph{Secondly}, the agent learns to build human-like deck. 
The human player says that: ``There is one match against AI's Flamewaker Mage. I'm killed when I have 21 hp. It's amazing.'' 
However, there is no prior \emph{domain knowledge} or \emph{expert data} during the training process and the models learn to build deck absolutely from-scratch, 
but the human player regards the AI's deck as Flamewaker Mage which is a popular choice in the \emph{Hearthstone} community. 
In Sec.\ref{sec:decks} we show these two decks.
\emph{Thirdly}, training time matters. 
Though c5-2day-vs-b4-2day has winrate $55\%$ in Tab. \ref{tab:baselines}, c5-16day performs worse than b4-23day when playing with human players. 
For instance, c5-16day uses Antique Healbot at full health at one match, which is an obvious mistake. 
But this has never happened on b4-23day. 
\emph{Fourthly}, cheat-on-deck can generate specific strategy against the human player. 
There is one match the AI takes two Antique Healbots that well limit the human player's Freeze Mage. 
The human player says that: ``AI takes so many healbots. It's almost killed and two healbots heal it!''.  
\emph{Fifthly}, cheat-on-deck encourages diversity. 
There is one match that the human player says that: ``The AI's warrior has Kel'Thuzad. It's unexpected.''

Our machine-vs-human evaluation shows that the combination of DRL and OSFP with carefully finetuned techniques is able to produce a master-level agent on card games. 
The learned agent is also human-like without any prior knowledge of the game, except that the observation space and the action space of the agent are carefully aligned with human players'. 
Cheat leads to adversarial policies against the opponent's.

\section{Conclusion}
\label{sec:conclusion}

In this work, we follow existed work on strategy card games and apply the algorithms to the famous game \emph{Hearthstone}. 
We further improve several techniques, which leads to better performance on \emph{Hearthstone}. 
We also do machine-vs-human evaluation, where our agents show master-level ability and human-like strategy. 
Our work show that the algorithm framework of combining E2E policy and OSFP is a promising way to train a master-level agent from scratch on strategy card games.

\subsection*{Acknowledgements}
Our colleagues Ze Chen and Wei Xi also contributed to the development and maintenance of the web GUI.
We thank the CoG 2023 anonymous reviewers for their helpful feedback that improves the quality of this paper.

\bibliography{main-cog2023}
\bibliographystyle{IEEEtran}

\section{Appendix}
\subsubsection{Hyperparameters}
\label{app:hp}
We follow most hyperparameters of \cite{xi2023master}, but a few are adjusted. The adjustments are highlighted in color. 

\begin{table}[htbp]
    \centering
    \begin{tabular}{c|c}
        \toprule
         Parameter & Value \\
         \midrule
         {\color{blue}Weight of policy gradient from PPO} & 1.0 \\
         Weight of policy gradient from UPGO & 1.0 \\
         Weight of value function loss & 1.0 \\
         Weight of entropy penalty & 0.01 \\
         Learning rate & 7e-5 \\
         Batch size & 1e+4 * 8gpu \\
         {\color{blue}Discount} & {\color{blue}1.0} \\
         LSTM states & 256 \\
         Sample reuse & 2 \\
         {\color{blue}V-Trace $c$ clip} & {\color{blue}[0.001, 1.007]} \\
         {\color{blue}V-Trace $\rho$ clip} & {\color{blue}[0.001, 1.007]} \\
         \bottomrule
    \end{tabular}
    \caption{Reinforcement learning hyperparameters. }
    \label{tab:rl_hyper_parameter}
\end{table}

\begin{table}[htbp]
    \centering
    \begin{tabular}{c|c}
        \toprule
         Parameter & Value \\
         \midrule
         Self-play Probability, $p$ & 0.6 \\
         {\color{blue}Add to historical model threshold, $\xi$} & {\color{blue}0.55} \\
         Add to historical model max LP, $c$ & 6 \\
         Num samples of each LP & 3.2e+8 \\
         \bottomrule
    \end{tabular}
    \caption{OSFP hyperparameters. The notations are corresponding to Alg. \ref{alg:spfp}. }
    \label{tab:osfp_hyper_parameter}
\end{table}


\subsubsection{Observation Space and Action Space}
\label{app:obs_and_act}
The observation space and action space are detailed in Tab. \ref{tab:obs}.
\begin{table}[htbp]
\begin{tabular}{l|l}
\toprule
Observations & Descriptions \\
\midrule
hero                   & my hero, one of (mage, hunter, warrior) \\
card set               & all cards, including each hero's specific cards \\
card selected mask     & 1 if the card has been selected else 0 \\
card can selected mask & 1 if the card can be selected else 0 \\
\midrule
hero           & my hero, one of (mage, hunter, warrior) \\
oppo hero      & opponent's hero, one of (mage, hunter, warrior) \\
my deck        & cards in my deck \\
decision type  & one of (construct, select, minion battlecry,  \\
               & spell card, minion/hero attack, hero power, \\
               & end turn) \\
my board       & minions on my board and their scalar features \\
oppo board     & minions on opponent's board and \\
               & their scalar features \\
my hand        & cards in my hand and their scalar features \\
my graveyard   & cards in my graveyard and their scalar features \\
oppo graveyard & cards in opponent's graveyard and \\
               & their scalar features \\
my player      & my features, such as number of hands and \\
               & minions, mana, weapon \\
oppo player    & opponent's features, such as number of hands \\
               & and minions, mana, weapon \\
BT action mask & 1 if the action can be done else 0 \\
\bottomrule
\end{tabular}
\caption{Observation Space. The top block is for CB and the bottom block is for BT. }
\label{tab:obs}
\end{table}

\begin{table}[htbp]
\begin{center}
\scalebox{0.85}{
\begin{tabular}{l|l}
\toprule
Action & Description \\
\midrule
selected card & card that is selected into the deck, corresponding to \\
              & \emph{card set} in observation space \\
\midrule
              & card that is selected as one of \emph{(type, target)}, one step for \emph{type} and \\
              & one step for \emph{target}, \\
selected card & \emph{type} $\in$ \emph{\{my hand card, my board card, opponent’s board card,} \\
              & \emph{my hero power card, end turn card\}}, \\
              & \emph{target} $\in$ \emph{\{my hero, opponent's hero, my board card and} \\
              & \emph{opponent's board card\}} \\
\bottomrule
\end{tabular}
}
\caption{Action Space. The top line is for CB and the bottom line is for BT. }
\label{tab:act}
\end{center}
\end{table}

\subsubsection{End-to-End Network}
\label{app:network}
\begin{figure}[htbp]
\centering
\includegraphics[width=1.0\linewidth]{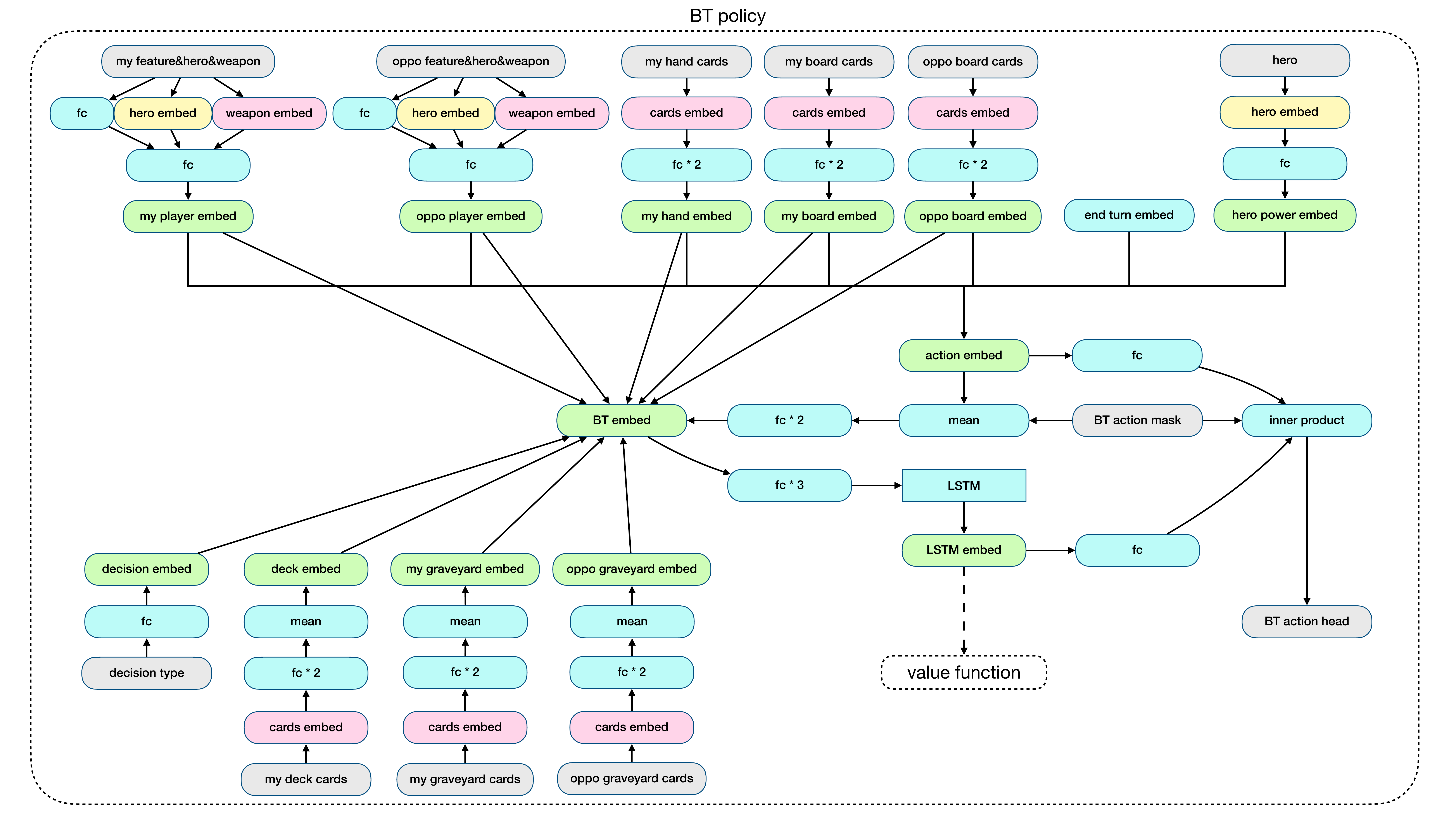}
\centering
\includegraphics[width=1.0\linewidth]{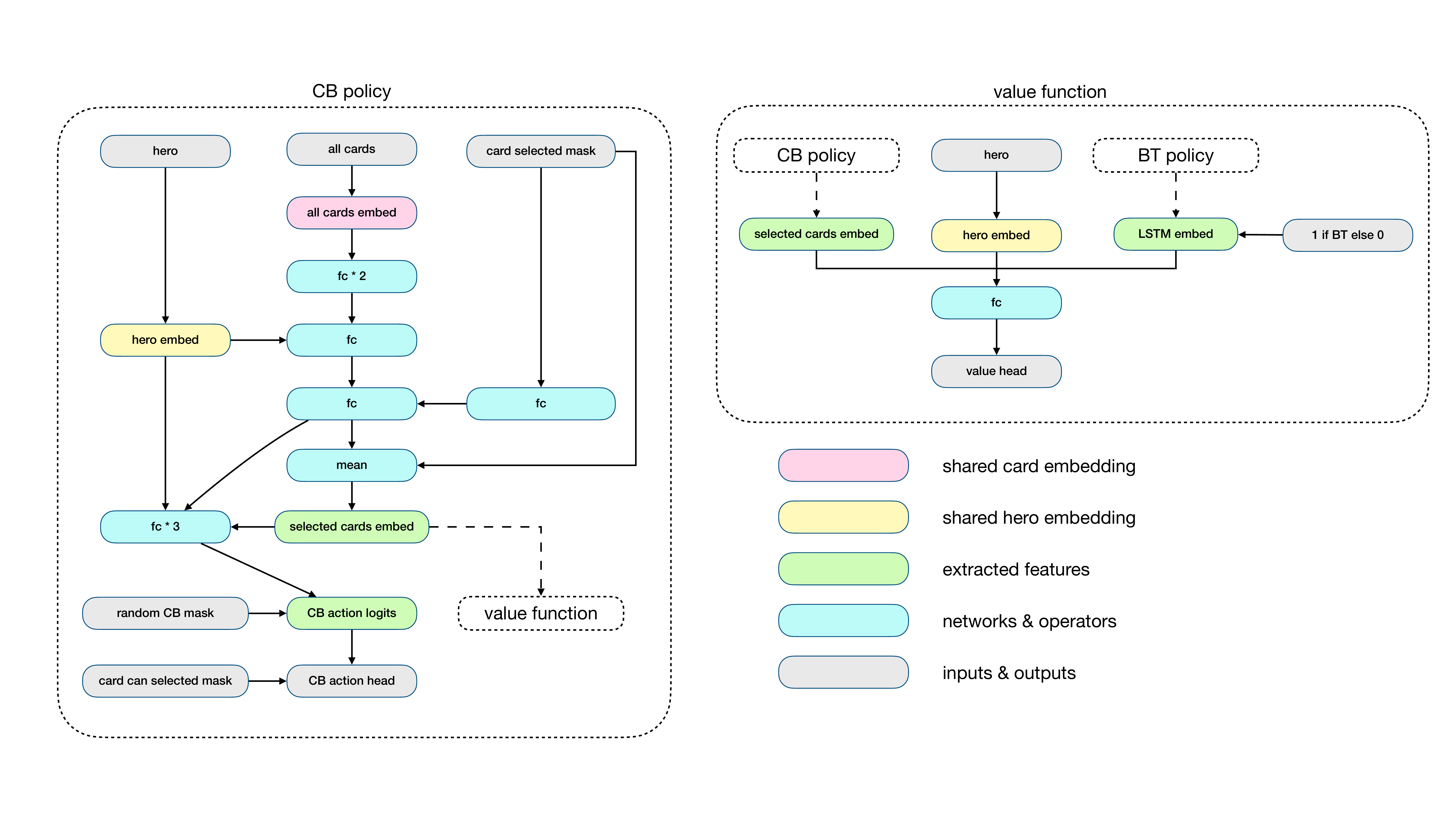}
\caption{Neural Network Structure. }
\label{fig:nn_structure}
\end{figure}

The E2E neural net architecture is given in Fig. \ref{fig:nn_structure}.


\subsubsection{Optimistic Smooth Fictitious Play}
\label{app:osfp}
Our optimistic smooth fictitious play strategy follows \cite{xi2023master}. For completeness, we describe it here. 
We define a \emph{learning period} (LP) to be a fixed number of training samples.
During each LP, we sample opponents according to the payoff table between current policy and historical policies. 
After each LP, we decide whether current policy should be added to historical policies. 
The complete codes are shown in Alg. \ref{alg:spfp} and the corresponding hyperparameters are listed in Tab. \ref{tab:osfp_hyper_parameter}. 

\begin{center}
\begin{algorithm}[htbp]
  \caption{Optimistic Smooth Fictitious Play. }  
      \begin{algorithmic}
        \STATE Init $H = []$.
        \STATE Init $0 < p < 1, 0 < \xi < 1$. 
        \STATE Init $c > 0$, $count = 0$.
        \STATE Init probability function $f$. 
        \FOR{$LP=0,1,2,\dots$}
            \STATE Init $G = C = [0] \text{ * len}(H)$. 
            \WHILE{LP not end}
                \FOR{each actor}
                    \IF{len($H$) = 0 \OR $\text{Unif}(0, 1) < p$}
                        \STATE Opponent player = current learner.
                        \STATE Finish game. 
                    \ELSE
                        \STATE Sample $i \sim f(\dots, (G[i], C[i]), \dots)$.
                        \STATE Opponent Player = $H[i]$.
                        \STATE Finish game and get $g \in \{+1, -1\}$.
                        \STATE $G[i] = G[i] + g$, $C[i] = C[i] + 1$. 
                    \ENDIF
                \ENDFOR
            \ENDWHILE
            \IF{($G[i] / C[i] > \xi, \forall\,i$) \OR ($count > c$)}
                \STATE Add current learner to $H$. 
                \STATE $count = 0$. 
            \ELSE
                \STATE $count = count + 1$. 
            \ENDIF
        \ENDFOR
      \end{algorithmic}
    \label{alg:spfp}
\end{algorithm}
\end{center}

\subsubsection{Compute Resource}
We rely on distributed RL training for all the experiments reported in this study.
To do so, we use an internal framework which is essentially an Actor-Learner architecture \cite{espeholt2018impala} that requires both GPUs and CPU cores for a single RL training.
For the best non-cheat training b4 reported in Tab. \ref{tab:baselines},
we employ 24 GPUs (V100) and 5856 CPU cores.

\subsubsection{Web GUI}
\label{app:gui}
\begin{figure}[htbp]
\centering
\includegraphics[width=1.0\linewidth]{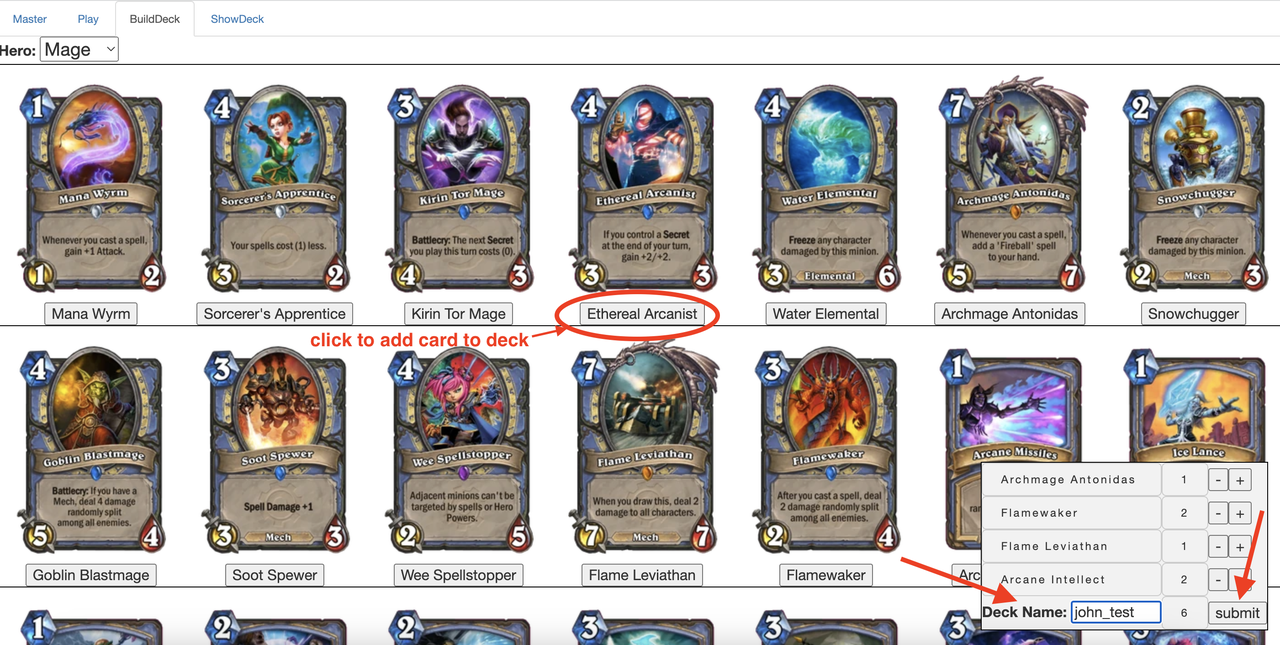}
\includegraphics[width=1.0\linewidth]{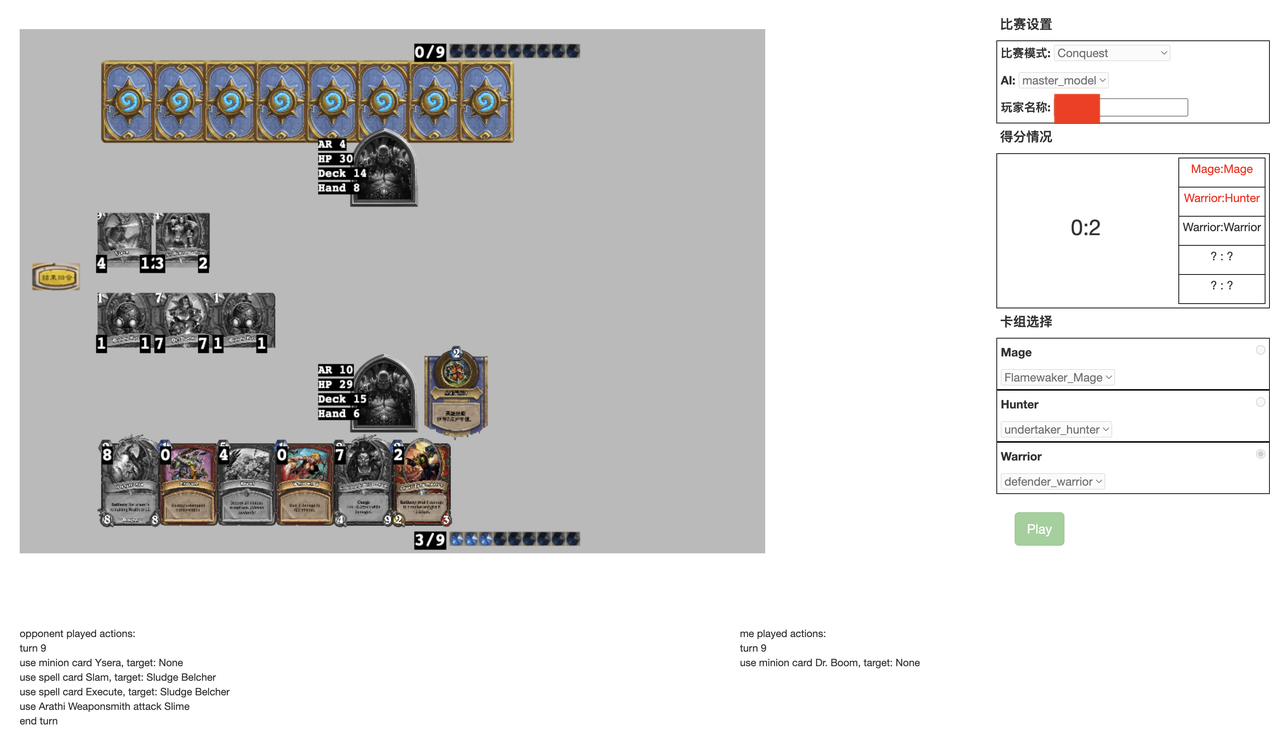}
\caption{Web GUI. 
Upper: deck building. 
The human player can browse all cards available to a particular hero (Mage, Warrior or Hunter) and select the 30 cards. 
The deck can be saved, and be retrieved in the later battle stage.
Lower: battle. 
The screen top is AI and the bottom is human player. 
The actions in previous turn taken by AI (denoted as ``opponent player'') and the human player (denoted as ``me'') are displayed as texts in the bottom screen.
}
\label{fig:gui}
\end{figure}

For the machine-vs-human test, 
we develop a dedicated web GUI where we can deploy models that play against a human player in remote, 
see Fig. \ref{fig:gui}. 
The human players can build their own decks and save, 
and play against AI models in a tournament.

\subsubsection{Decks}
\label{sec:decks}

\begin{figure}[htbp]
\centering
\subfigure[]{
    \label{fig:deck-ai}
    \includegraphics[width=1.0\linewidth]{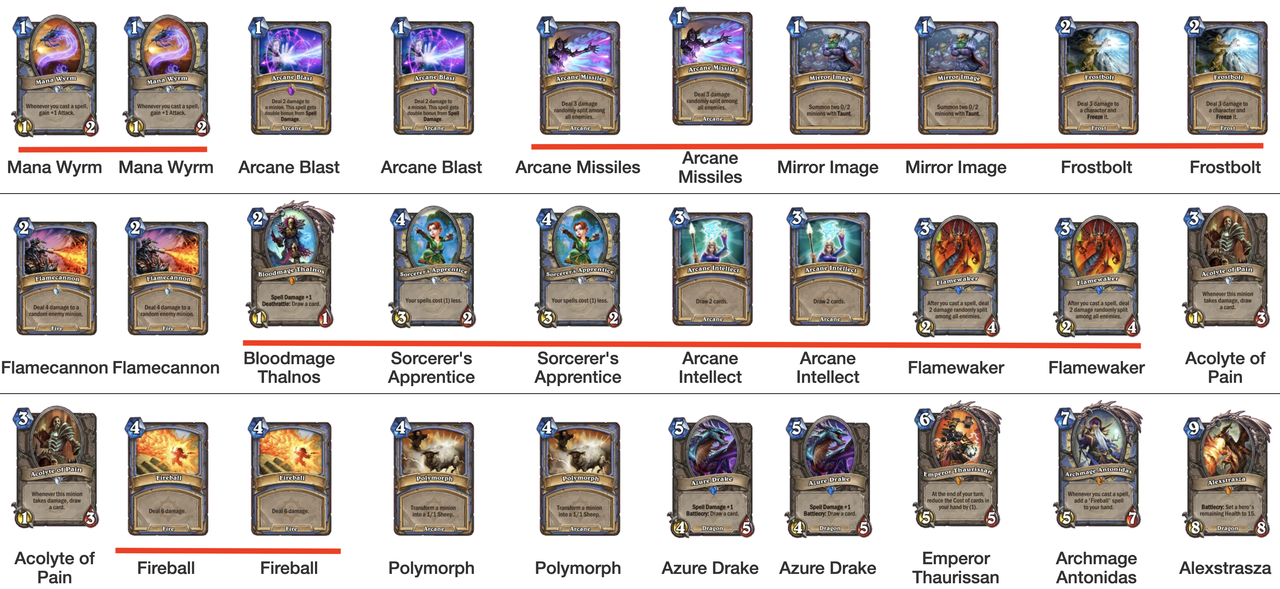}
}
\centering
\subfigure[]{
    \label{fig:deck-human}
    \includegraphics[width=1.0\linewidth]{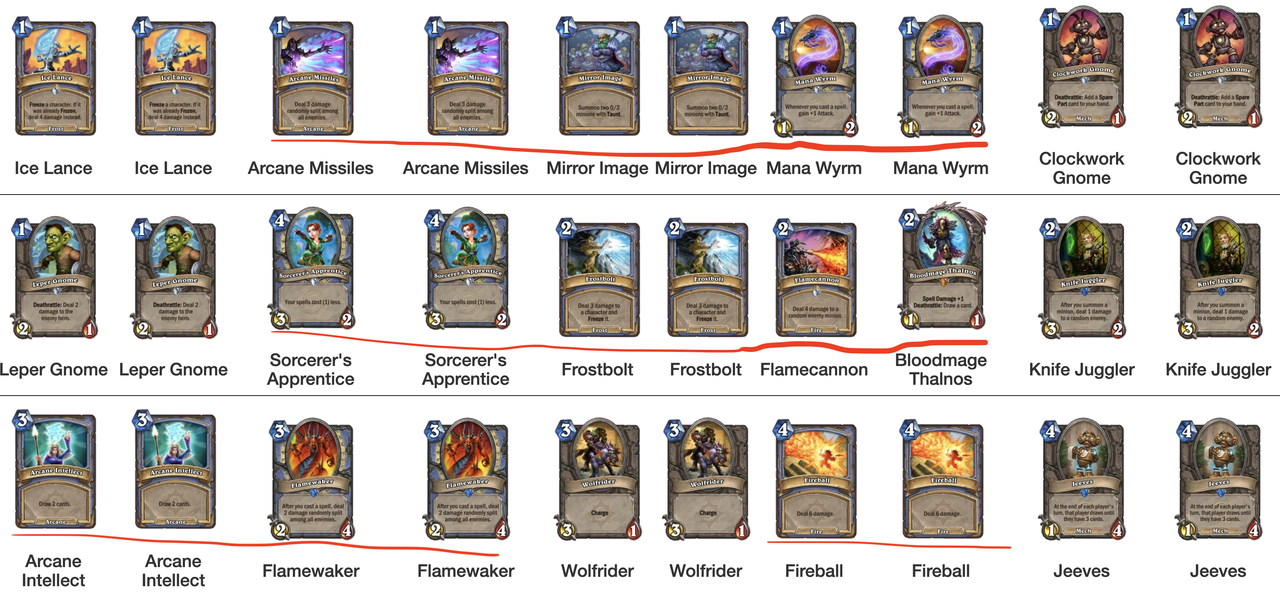}
}
\caption{
The complete 30-card decks. 
(a): a Mage deck learnt by AI; 
(b): the human deck ``Flamewaker Mage'' that was popular in around April 2015.
}
\label{fig:deck}
\end{figure}

As mentioned in Sec.\ref{sec:random_cb} and Sec.\ref{sec:machine_vs_human}, 
we provide the AI deck and the human deck in Fig.\ref{fig:deck}.

\subsubsection{Game-Play Videos}
\label{sec:video}

We provide videos of two selected matches in the human-vs-AI evaluation, 
named as ``mmt1''\footnote{\url{https://youtu.be/ht_EZbMJjmA}} and ``mmt2''\footnote{\url{https://youtu.be/YnW2ypZL4HY}}, respectively.
In both videos, 
top screen is AI player with the b4-23day model (best normal version), 
bottom screen is human player who is the Hearthstone streamer, 
as described in Sec.V of the main texts. 

The videos are made by checking the replay files and recording the game-play, 
therefore the hand cards and secrets of both players are visible to the viewer. 
During the human-vs-AI test, however, such information is hidden from each other player, 
as in the real Hearthstone games.

Here are the brief descriptions of the two videos.

``mmt1''. 
AI uses the learnt deck which is much like the ``Flamewaker Mage'' in real world (See main texts),
while the human player uses the so-called ``Freeze Mage'' deck.
Early in the match, 
both AI and human player choose to draw more cards by using the effect of the cards from their deck, 
intending for gaining the upper hand.
In the mid-match,
both players place the card ``Emperor Thaurissan'' to reduce mana; 
However, AI manages to accumulate more advantages by successfully playing several combos, 
and preserves more hand cards during the course.
Late in the match, 
the human player still has 21 Hit Points (out of the full 30 HPs), 
and henceforth does not choose to place the secret ``Ice Block'' by considering that 21 HPs should be safe. 
On the other hand, 
AI has, however, obtained the card ``Flamewaker'' and many other spell cards by drawing sufficient cards in the early- to mid-match,
resulting to a potential damage that reaches 31 HPs. 
Consequently, AI performs a One-Turn-Kill (OTK) that ends the match. 
AI seems to adopt a typical ``accumulate-hand-cards-and-outbreak'' strategy.

``mmt2''.
In regards of deck, 
this match is still the AI's learnt Flamewaker Mage vs the human player's Freeze Mage.
In this match, 
AI constantly pushes the human player by maximizing the mid- to short-term benefit at each turn and gaining the upper hand gradually.
Finally, AI wins the match. 
Compared to ``mmt1'', 
the strategy that AI adopts in this match seems more deliberate and less risky.




\end{document}